\documentclass[journal]{IEEEtran}
\usepackage{amsmath,amsfonts}
\usepackage{algorithmic}
\usepackage{algorithm}
\usepackage{array}
\usepackage[caption=false,font=normalsize,labelfont=sf,textfont=sf]{subfig}
\usepackage{textcomp}
\usepackage{stfloats}
\usepackage{url}
\usepackage{verbatim}
\usepackage{graphicx} 
\usepackage{cite}
\usepackage{multirow}
\usepackage{booktabs}
\usepackage{color}
\usepackage{balance}
\usepackage[table]{xcolor}
\begin{document}

\title{Leveraging the Powerful Attention \\of a Pre-trained Diffusion Model \\for Exemplar-based Image Colorization}

\author{Satoshi Kosugi
\thanks{The author is with the Institute of Integrated Research, Institute of Science Tokyo, Yokohama, Kanagawa 226-8503, Japan 
(e-mail: kosugi.s.2cf1@m.isct.ac.jp)

Copyright © 2025 IEEE. Personal use of this material is permitted. However, permission to use this material for any other purposes must be obtained from the IEEE by sending an email to pubs-permissions@ieee.org.
}
}

\markboth{IEEE TRANSACTIONS ON CIRCUITS AND SYSTEMS FOR VIDEO TECHNOLOGY, VOL. XX, NO. XX, 2025}
{Satoshi Kosugi: Leveraging the Powerful Attention of a Pre-trained Diffusion Model for Exemplar-based Image Colorization}

\maketitle

\begin{abstract}
  Exemplar-based image colorization aims to colorize a grayscale image using a reference color image,
  ensuring that reference colors are applied to corresponding input regions based on their semantic similarity.
  To achieve accurate semantic matching between regions, we leverage the self-attention module of a pre-trained diffusion model, which is trained on a large dataset and exhibits powerful attention capabilities.
  To harness this power, we propose a novel, fine-tuning-free approach based on a pre-trained diffusion model, making two key contributions.
  First, we introduce dual attention-guided color transfer.
  We utilize the self-attention module to compute an attention map between the input and reference images, effectively capturing semantic correspondences.
  The color features from the reference image is then transferred to the semantically matching regions of the input image, guided by this attention map,
  and finally, the grayscale features are replaced with the corresponding color features.
  Notably, we utilize dual attention to calculate attention maps separately for the grayscale and color images, achieving more precise semantic alignment.
  Second, we propose classifier-free colorization guidance, which enhances the transferred colors by combining color-transferred and non-color-transferred outputs.
  This process improves the quality of colorization.
  Our experimental results demonstrate that our method outperforms existing techniques in terms of image quality and fidelity to the reference.
  Specifically, we use 335 input-reference pairs from previous research, achieving an FID of 95.27 (image quality) and an SI-FID of 5.51 (fidelity to the reference).
  In addition, we evaluate on our novel dataset, which consists of 100 pairs of natural photos and historical paintings, achieving an FID of 219.05 and an SI-FID of 7.94.
  Our source code is available at https://github.com/satoshi-kosugi/powerful-attention.
\end{abstract}


\section{Introduction}
\label{sec:intro}
\IEEEPARstart{W}{ith} the recent developments in image processing technology, there has been significant attention paid to research on the restoration of image quality, 
including image denoising~\cite{yue2019ienet,guo2023towards}, image deblurring~\cite{zhang2021deep,10264126}, image dehazing~\cite{kumar2021dynamic,kumar2021improved}, and image enhancement~\cite{kosugi2020unpaired,9792425,kosugi2023crowd,kosugi2024prompt,kosugi2023personalized}.
In this context, we focus on image colorization.
Colorizing grayscale images has long been a significant challenge in computer vision, with applications ranging from historical photo restoration to medical imaging enhancement and visual media improvement.
Traditional colorization techniques often struggle with ambiguity, as multiple color choices may be equally valid for a single image,
and the final output may not align with the user's intent.
To address this, we explore exemplar-based image colorization, where the goal is to colorize the grayscale input by applying colors from a reference image,
considering region-wise semantic correspondences.
For example, if the reference image contains a red flower and green leaves, we aim to colorize the corresponding regions in the input image with red for the flower and green for the leaves.

To better capture semantic correspondences, we leverage the self-attention of a pre-trained diffusion model.
The diffusion model generates images by progressively removing noise from a noisy input image through a denoising network, where the self-attention mechanism plays a critical role.
Recent diffusion models are trained on extensive datasets;
for instance, Stable Diffusion~\cite{rombach2022high} is trained on the LAION dataset~\cite{schuhmann2022laion}, which contains over five billion images.
Given that previous attention-based colorization methods~\cite{lu2020gray2colornet,yin2021yes,zhang2022scsnet,bai2022semantic,carrillo2022super,wang2023unsupervised,zou2024lightweight}
are trained using smaller datasets from scratch,
the self-attention of the pre-trained diffusion model is more powerful.
By harnessing this power, we can develop a colorization method that more accurately captures semantic correspondences.

In this paper, we introduce a novel, fine-tuning-free approach leveraging a pre-trained diffusion model.
Our method offers two key contributions.
First, we present {\bf 1) dual attention-guided color transfer}, where we modify self-attention, originally designed for single images, to function as cross-attention.
Using DDIM inversion~\cite{song2020denoising}, we transform both the input grayscale image and the reference color image into noisy images.
When denoising these noisy images, we compute an attention map between the input and reference images:
the query features are derived from the input image, while the key features are taken from the reference image, resulting in an attention map formed from these query and key features.
This attention map is then applied to the value features extracted from the reference image, yielding the output features.
In the attention map, higher scores typically represent semantically similar regions, and since the value features from the reference image carry color information,
this mechanism allows us to transfer colors to semantically corresponding regions in the input image.
Because directly calculating attention between the grayscale input image and the color reference image can lead to inaccurate semantic correspondences,
we utilize dual attention consisting of gray-to-gray and colorized-to-color attention, which leads to more precise semantic alignment.

We found that using only the dual attention-guided color transfer often results in unnatural colorization.
To achieve higher-quality results, we introduce our second contribution, {\bf 2) classifier-free colorization guidance},
which is inspired by the concept of classifier-free guidance~\cite{ho2021classifier}.
During the denoising process, we perform two forward passes through the denoising network: one with color transfer applied and one without.
We then extrapolate the color-transferred output using the non-color-transferred output.
This enhances the transferred color features, resulting in more natural colorization.

We conduct experiments to evaluate our method using input-reference pairs from existing research~\cite{he2018deep}.
Our method achieves superior performance compared to existing methods in terms of image quality and fidelity to the reference.
Visual comparisons demonstrate that it achieves more accurate semantic correspondences.
Furthermore, we conduct experiments using pairs of grayscale historical paintings and contemporary color photos,
demonstrating that our method effectively captures semantic correspondences even in this challenging scenario.

Our contributions are as follows:
\begin{itemize}
\vspace{1mm}
\item We propose dual attention-guided color transfer.
By transferring features from the reference image into the input image denoising process, with attention as a guide, we achieve semantically consistent colorization.
\vspace{1mm}
\item We propose classifier-free colorization guidance, which emphasizes the differences between outputs with and without the color transfer to
improve the quality.
\vspace{1mm}
\item Our method outperforms existing approaches by effectively utilizing a pre-trained diffusion model.
\end{itemize}

\section{Related Works}
\label{sec:relatedworks}

\subsection{Exemplar-based Colorization}

In exemplar-based colorization, computing semantic correspondences is crucial.
Existing methods can be broadly divided into two main categories from the perspective of semantic correspondence computation:
{\bf 1) predefined feature-based methods} and {\bf 2) attention-based methods}.

In 1) predefined feature-based methods, features are extracted from regions of both the input and reference images,
and the reference colors are transferred to regions with similar features.
Welsh et al.~\cite{welsh2002transferring} were the first to propose an exemplar-based colorization method,
which calculates region-wise correspondences between the input and reference images based on luminance distribution.
Morimoto et al.~\cite{morimoto2009automatic} improved upon the method of Welsh et al.~\cite{welsh2002transferring}
and proposed a technique for automatically selecting reference images using the GIST scene descriptor~\cite{oliva2001modeling}.
Various handcrafted features have been used to compute the region-wise correspondences:
Ironi et al.~\cite{ironi2005colorization} employed discrete cosine transform coefficients,
Charpiat et al.~\cite{charpiat2008automatic} and Gupta et al.~\cite{gupta2012image} utilized the SURF descriptor~\cite{bay2006surf},
Chia et al.~\cite{chia2011semantic} and Liu et al.~\cite{liu2008intrinsic} relied on SIFT features~\cite{lowe1999object},
and Fang et al.~\cite{fang2019superpixel} used both HOG~\cite{dalal2005histograms} and DAISY~\cite{tola2009daisy} descriptors.
To capture higher-level semantics, several studies~\cite{he2018deep,xu2020stylization,li2021globally,huang2022unicolor,leduc2024non} have utilized features extracted from VGG~\cite{simonyan2014very}, which was pre-trained on ImageNet~\cite{deng2009imagenet}.
Specifically, He et al.\cite{he2018deep} trained VGG-19 using grayscale images to achieve accurate semantic correspondence between input and reference grayscale images.
Huang et al.\cite{huang2022unicolor} proposed a method that accepts not only reference images but also strokes and text as conditioning inputs.
Leduc et al.~\cite{leduc2024non} aggregated VGG-19 features based on superpixels.
A key advantage of these methods is their ability to leverage a variety of predefined features.
However, a major drawback is that these features are fixed and not optimized for semantic correspondence, resulting in limited accuracy of semantic correspondence.

For better semantic correspondence, 2) attention-based methods~\cite{lu2020gray2colornet,yin2021yes,zhang2022scsnet,bai2022semantic,carrillo2022super,wang2023unsupervised,zou2024lightweight,liang2024control} have been proposed.
An attention map between the input and reference features is computed, and the reference colors are transferred to the input guided by this attention map.
Various improvements have been introduced to enhance attention mechanisms.
Lu et al.~\cite{lu2020gray2colornet} introduced a gating mechanism to efficiently combine colors from the reference image with prior knowledge.
Yin et al.~\cite{yin2021yes} developed a unified framework that considers color information from both a reference image and a database simultaneously.
Zhang et al.~\cite{zhang2022scsnet} proposed a framework capable of simultaneous super-resolution and colorization.
Bai et al.~\cite{bai2022semantic} improved the semantic correspondence using a sparse attention mechanism.
Carrillo et al.~\cite{carrillo2022super} reduced computational complexity by utilizing superpixel-based features.
Wang et al.~\cite{wang2023unsupervised} achieved more detailed semantic correspondences through pyramid dual non-local attention.
Zou et al.~\cite{zou2024lightweight} introduced a semantic attention-guided laplacian pyramid to separate the foreground and background in semantic correspondences.
A key advantage of these methods is that attention-based methods optimize both feature extraction and feature matching mechanisms in an end-to-end manner, enabling more accurate semantic correspondence computation.
However, a major drawback is that training attention modules is computationally expensive, and without a sufficiently large dataset, it is difficult to fully exploit their potential.

In this paper, we propose a novel attention-based method that leverages the attention mechanisms of a pre-trained diffusion model.
A key advantage of our approach is that it benefits from large-scale data and fully utilizes the potential of attention modules, leading to more accurate semantic correspondences.

\subsection{Attention-based Image Editing with Pre-trained Diffusion Models}
Diffusion models have achieved significant success in image generation, with the attention mechanism playing a crucial role.
Various fine-tuning-free image editing methods have been proposed to leverage the attention mechanism.
Hertz et al.~\cite{hertz2022prompt} introduced a text-based image editing method,
discovering that cross-attention between text and an image significantly contributes to the structure of a generated image.
Tumanyan et al.~\cite{tumanyan2023plug} found that spatial features and self-attention maps are crucial for shaping the image structure,
and they inject these features into the target image generation process.
Cao et al.~\cite{cao2023masactrl} proposed mutual self-attention control,
where the target key and value features are replaced with those of the source image.
Chung et al.~\cite{chung2024style} achieved style transfer by substituting the key and value features of the content image with those from the style image.
Liu et al.~\cite{liu2024towards} conducted an in-depth analysis of cross- and self-attention,
concluding that self-attention plays a greater role in preserving image structure.
Following these analyses,
we aim to harness the attention capabilities of pre-trained diffusion models for exemplar-based colorization.

\begin{figure*}[t]
  \centering
  \includegraphics[width=0.9\linewidth]{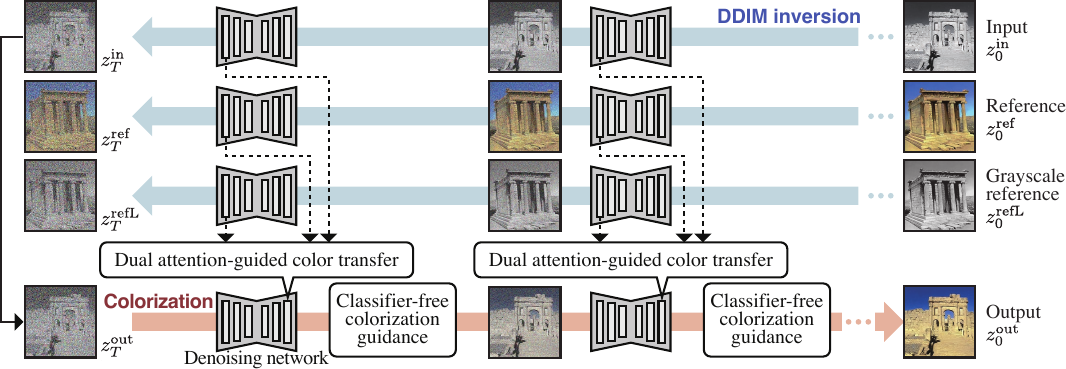}
  \caption{{Overview of our method.}
  First, $z_0^{\rm in}$, $z_0^{\rm ref}$, and $z_0^{\rm refL}$ are given; these are the latent variables corresponding to the input, reference, and grayscale reference, respectively.
  We apply DDIM inversion to these latent variables to obtain $z_T^{\rm in}$, $z_T^{\rm ref}$, and $z_T^{\rm refL}$.
  Next, we set $z_T^{\rm in}$ as the initial value for $z_T^{\rm out}$ and apply the colorization process to $z_T^{\rm out}$. This process consists of two key components:
  1) dual attention-guided color transfer and 2) classifier-free colorization guidance.
  Finally, we obtain the colorized output $z_0^{\rm out}$.
  }
  \label{fig1}
\end{figure*}

\section{Preliminary}
\label{sec:preliminary}
In this section, we introduce the diffusion model, which plays a crucial role in our research.
The purpose of a diffusion model is to generate realistic images by progressively denoising random noise.
In this paper, we use Stable Diffusion~\cite{rombach2022high}, where an encoder ${\rm E()}$ is applied to the image $x\in{\mathbb R}^{H\times W\times 3}$,
and the diffusion process is applied to the latent representation $z = {\rm E}(x)$, where $z\in{\mathbb R}^{h\times w\times c}$.
Here, $H$ and $W$ denote the height and width of the image, while $h$, $w$, and $c$ denote the height, width, and number of channels of the latent representation, respectively.
When generating a new image, random noise $z_{T_{\text{max}}}$ is progressively denoised to obtain $z_0$,
and an image is reconstructed using a decoder as $x = \mathrm{D}(z_0)$.

For denoising random noise, a denoising model $\varepsilon_{\theta}()$ is used.
$\varepsilon_{\theta}()$ is trained to predict noise as follows:
\begin{equation}
L_{\rm LDM} = \mathbb{E}_{z_0, \varepsilon, t} \left\| \varepsilon - \varepsilon_{\theta}(z_t, t, \mathcal{C}) \right\|_2^2.
\end{equation}
Here, $\varepsilon, \varepsilon_{\theta}(z_t, t, \mathcal{C}) \in \mathbb{R}^{h \times w \times c}$.
$\varepsilon$ represents Gaussian noise, where each element is independently sampled from $\mathcal{N}(0, 1)$.
$t \in \{1, \ldots, T_{\text{max}}\}$ denotes the time step, and $\mathcal{C}$ represents the text condition.

Stable Diffusion is capable of generating images without any text condition. In such cases, an empty string (``'') is typically used as the text condition.
In our work, we do not utilize any text condition either, and thus we provide an empty string to the model.
For simplicity, we denote $\varepsilon_{\theta}(z_t, t, \mathcal{C})$ as $\varepsilon_{\theta}(z_t)$.

\vspace{2mm}
\noindent
{\bf DDIM sampling.}~
To generate images, we use DDIM sampling~\cite{song2020denoising}. Denoising\>processes\>are\>iteratively\>applied\>as
\begingroup
\setlength{\jot}{10pt}
\begin{equation}
\begin{split}
&z_{t-1} = c_1 z_t + c_2  \varepsilon_\theta(z_t),~{\rm where}\\
c_1 = \sqrt{\frac{\alpha_{t-1}}{\alpha_t}},~&c_2 = \sqrt{\alpha_{t-1}} \left( \sqrt{\frac{1}{\alpha_{t-1}} - 1} - \sqrt{\frac{1}{\alpha_t} - 1} \right).
\nonumber
\end{split}
\end{equation}
\endgroup
$c_1, c_2$, and $\alpha_t$ are scalar values.
Details about $\alpha_t$ can be found in \cite{song2020denoising}.

\vspace{2mm}
\noindent
{\bf DDIM inversion.}~
DDIM inversion~\cite{song2020denoising} is used to find the noise $z_{T_{\text{max}}}$ that corresponds to the real data $z_0$. This process is formulated as follows.
\begingroup
\setlength{\jot}{10pt}
\begin{equation}
\begin{split}
&z_{t+1} = c'_1 z_t + c'_2  \varepsilon_\theta(z_t),~{\rm where}\\
c'_1 = \sqrt{\frac{\alpha_{t+1}}{\alpha_t}},~&c'_2 = \sqrt{\alpha_{t+1}} \left( \sqrt{\frac{1}{\alpha_{t+1}} - 1} - \sqrt{\frac{1}{\alpha_t} - 1} \right).
\nonumber
\end{split}
\end{equation}
\endgroup

\vspace{2mm}
\noindent
{\bf Self-attention.}~
Self-attention modules play an important role in the denoising model $\varepsilon_{\theta}()$.
The model $\varepsilon_{\theta}()$ adopts a U-Net architecture~\cite{ronneberger2015u},  that incorporates several self-attention layers.
We denote the input feature to a self-attention layer as $\phi \in \mathbb{R}^{M \times d}$, where $M$ is the number of flattened spatial positions and $d$ is the feature dimension.
The input $\phi$ is first projected into query, key, and value features using learnable linear transformations:
\begin{equation} 
  q = \phi W_q~, \quad k = \phi W_k~, \quad v = \phi W_v~,
\end{equation} 
where $q, k, v \in \mathbb{R}^{M \times d}$ are the query, key, and value features, and $W_q, W_k, W_v \in \mathbb{R}^{d \times d}$ are the corresponding projection matrices.
Next, a similarity matrix $S \in \mathbb{R}^{M \times M}$ is computed via scaled dot-product attention:
\begin{equation}
  S = \frac{q k^\top}{\sqrt{d}~}. 
\end{equation} 
The attention map $A \in \mathbb{R}^{M \times M}$ is then obtained by applying the softmax function row-wise:
\begin{equation}
  A = \mathrm{softmax}(S).
\end{equation}
Finally, the output feature $\overline{\phi} \in \mathbb{R}^{M \times d}$ is computed as a weighted aggregation of the value features:
\begin{equation}
  \overline{\phi} = A v.
\end{equation}

Our method is primarily based on Stable Diffusion. Further technical details can be found in the original paper on Stable Diffusion~\cite{rombach2022high}.

\section{Proposed Method}
\label{sec:proposedmethod}
Given a grayscale input image $x^{\rm in} \in \mathbb{R}^{H \times W \times 1}$ and a color reference image $x^{\rm ref} \in \mathbb{R}^{H \times W \times 3}$,
our goal is to transfer the colors from $x^{\rm ref}$ to $x^{\rm in}$, considering semantic correspondence.
The output image is represented as $x^{\rm out}$. We propose a fine-tuning-free colorization method that leverages a pre-trained diffusion model.

We present an overview of our method in Fig.~\ref{fig1}.
First, we extract the luminance from the reference color image $ x^{\rm ref}$
to create a grayscale image $ x^{\rm refL} \in \mathbb{R}^{H \times W \times 1} $.
Specifically, we convert $x^{\rm ref}$ to the Lab color space~\cite{connolly1997study}, and define its L channel as $x^{\rm refL}$.
The images $x^{\rm in}$, $x^{\rm ref}$, and $x^{\rm refL} $ are then converted into latent variables using an encoder, which we denote as $ z_0^{\rm in}$, $z_0^{\rm ref}$, and $z_0^{\rm refL} $, respectively.
Using DDIM inversion~\cite{song2020denoising}, we transform  $z_0^{\rm in}$, $z_0^{\rm ref}$, and $z_0^{\rm refL}$ into $z_T^{\rm in}$, $z_T^{\rm ref}$, and $z_T^{\rm refL}$.
We first define $ z_T^{\rm out} $ as $ z_T^{\rm in} $, {\it i.e.},
\begin{equation}
z_T^{\rm out} \leftarrow z_T^{\rm in}.
\end{equation}
Then, we apply DDIM sampling iteratively to $ z_T^{\rm out} $ as
\begin{equation}
z_{t-1}^{\rm out} = c_1 z_t^{\rm out} + c_2 \tilde{\varepsilon}_\theta(z_t^{\rm out}).
\end{equation}
Here, the denoising model $ \tilde{\varepsilon}_\theta() $ incorporates our proposed method, which transfers colors from the reference image.
Finally, applying the decoder to the resulting  $z_0^{\rm out}$ produces the output  $x^{\rm out} = {\rm D}(z_0^{\rm out})$.
Our method includes two major contributions: 1) dual attention-guided color transfer and 2) classifier-free colorization guidance, both of which we describe in detail below.

\subsection{Dual Attention-guided Color Transfer}
We propose dual attention-guided color transfer to apply the reference colors to semantically corresponding regions of an input image.
We leverage the self-attention of a pre-trained diffusion model in a manner similar to cross-attention.
Specifically, the query features are computed from the input image, while the key features are computed from the reference image, with an attention map derived from these components.
The attention map is applied to the value features computed from the reference image.
In the attention map, higher scores typically represent semantically similar regions.
Since the value features from the reference image carry color information, this mechanism allows the transfer of colors to semantically corresponding regions in the input image.

\begin{figure}[t]
    \centering
    \includegraphics[width=0.9\linewidth]{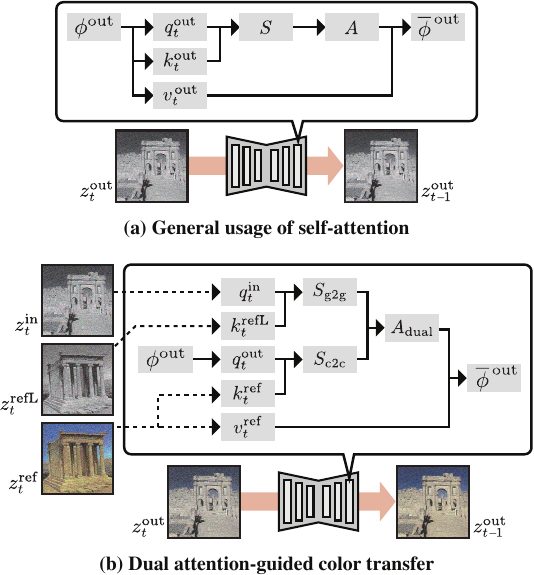}
    \caption{{Illustration of (a) general usage of self-attention~\cite{rombach2022high} and (b) dual attention-guided color transfer.}}
    \label{fig2}
\end{figure}

The input is a grayscale image, while the reference is a color image.
As a result, directly computing attention between them fails to establish appropriate semantic correspondences.
To address this, we propose using dual attention mechanisms: one for {\bf 1) gray-to-gray attention} and another for {\bf 2) colorized-to-color attention}.
We illustrate these mechanisms in Fig.~\ref{fig2}(b).
For comparison, we also show the general usage of self-attention~\cite{rombach2022high} in Fig.~\ref{fig2}(a).

In 1) gray-to-gray attention, we calculate the similarity map using the input image $z_t^{\rm in}$ and the grayscale reference image $z_t^{\rm refL}$.
During DDIM inversion, we collect the query features $q_t^{\rm in}$ from $z_t^{\rm in}$ and the key features $k_t^{\rm refL}$ from $z_t^{\rm refL}$ at each time step.
A gray-to-gray similarity map $S_{\rm g2g}$ is then computed using these features as follows:
\begin{equation}
  S_{\rm g2g} = \frac{q_t^{\rm in}(k_t^{\rm refL})^T}{\sqrt{d}}.
\end{equation}

In 2) colorized-to-color attention, the similarity map is computed using the colorized output feature $z_t^{\rm out}$ and the reference color image $z_t^{\rm ref}$. 
The query feature $q_t^{\rm out}$ is calculated from the feature $\phi^{\rm out}$ of the output $z_t^{\rm out}$ as $q_t^{\rm out} = \phi^{\rm out}W_q$. 
During DDIM inversion, key features $k_t^{\rm ref}$ are collected from $z_t^{\rm ref}$ at each time step.
A colorized-to-color similarity map $S_{\rm c2c}$ is then computed as follows:
\begin{equation}
S_{\rm c2c} = \frac{q_t^{\rm out}(k_t^{\rm ref})^T}{\sqrt{d}}.
\end{equation}

By combining gray-to-gray and colorized-to-color attention, the dual attention map $A_{\rm dual}$ is obtained:
\begin{equation}\label{dual_attention}
A_{\rm dual} = {\rm softmax}(S_{\rm g2g} \times \gamma + S_{\rm c2c} \times (1 - \gamma)),
\end{equation}
where $\gamma$ is a hyperparameter.
Using this attention map as guidance, the value features $v_t^{\rm ref}$ of the reference image $z_t^{\rm ref}$ are transferred as follows:
\begin{equation}\label{output}
\overline{\phi}^{\rm\,out} = A_{\rm dual}v_t^{\rm ref}.
\end{equation}
Since the attention map $A_{\rm dual}$ serves as a semantic guide,
the reference color information in the value features can be transferred to the semantically corresponding regions of the input image.

\begin{figure}[t]
    \centering
    \includegraphics[width=0.9\linewidth]{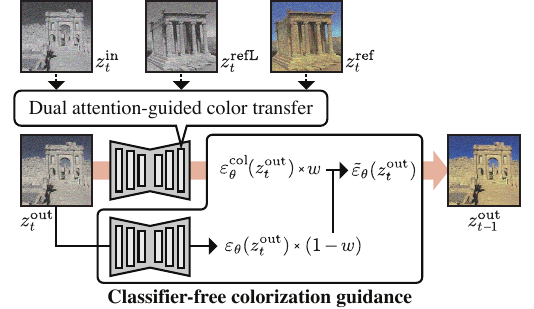}
    \caption{{Illustration of classifier-free colorization guidance.}}
    \label{fig3}
\end{figure}

\vspace{2mm}
\noindent
{\bf Self-attention injection.}~
In previous research~\cite{tumanyan2023plug}, it has been shown that injecting self-attention from the original image helps preserve the image layout.
Following this approach, we inject self-attention.
During DDIM inversion, we collect the query features $q_t^{\rm in}$ and key features $k_t^{\rm in}$ from the input $z_t^{\rm in}$, and calculate the attention map as
\begin{equation}
  A_{\rm in} = {\rm softmax}\left(\frac{q_t^{\rm in}(k_t^{\rm in})^T}{\sqrt{d}}\right).
\end{equation}
Next, we define the value features computed from the output features $\phi^{\rm out}$ as $v_t^{\rm out} = \phi^{\rm out}W_v$, which are weighted using $A_{\rm in}$.
Finally, we modify Eq.~(\ref{output}) as follows:
\begin{equation}
  \overline{\phi}^{\rm\,out} = A_{\rm dual}v_t^{\rm ref} \times (1 - \beta) + A_{\rm in}v_t^{\rm out} \times \beta,
\end{equation}
where $\beta$ is a hyperparameter.

\subsection{Classifier-free Colorization Guidance}
We propose classifier-free colorization guidance to improve the colorization process.
This method is inspired by classifier-free guidance~\cite{ho2021classifier},
a technique that enhances the effectiveness of a text prompt's condition by comparing a conditional output with an unconditional output.
Building on this method, we introduce classifier-free colorization guidance, as illustrated in Fig.~\ref{fig3}.
Our approach involves running the denoising model twice: once with dual attention-guided color transfer and once without it ({\it i.e.}, using general self-attention).
The output generated with dual attention-guided color transfer is denoted as $\varepsilon^{\rm col}_\theta(z_t^{\rm out})$.
We then enhance the colorization process by incorporating the output without color transfer.
\begin{equation}\label{color_enhancement}
\tilde{\varepsilon}_\theta(z_t^{\rm out}) = \varepsilon^{\rm col}_\theta(z_t^{\rm out}) \times w + \varepsilon_\theta(z_t^{\rm out}) \times (1 - w).
\end{equation}
Here, $w$ is a hyperparameter.
By setting $w$ such that $w > 1$, the transferred color is amplified, resulting in enhanced colorization.

The rationale for setting $w > 1$ in Eq.~(\ref{color_enhancement}) may not be immediately clear. 
To clarify this, we can rewrite Eq.~(\ref{color_enhancement}) as follows.
\begin{equation}
 \tilde{\varepsilon}_\theta(z_t^{\rm out}) = \varepsilon^{\rm col}_\theta(z_t^{\rm out}) + (\varepsilon^{\rm col}_\theta(z_t^{\rm out}) - \varepsilon_\theta(z_t^{\rm out})) \times (w - 1).
\end{equation}
Here, 
$(\varepsilon^{\rm col}_\theta(z_t^{\rm out}) - \varepsilon_\theta(z_t^{\rm out}))$ represents the difference between the color-transferred output and the non-color-transferred output.
By setting $w$ such that $w > 1$, this difference can be amplified, thereby enhancing the transferred colors and resulting in more natural colorization.

\subsection{Other Details}
\noindent
{\bf Early stopping of DDIM inversion.}~
We stop the DDIM inversion process early by setting $T$ such that $T < T_{\rm max}$ to reduce computational costs.
This early stopping prevents the input data from being fully converted into noise.
However, our goal is colorization, and the input $z_0^{\rm in}$ and the colorized result $z_0^{\rm out}$ should maintain the same structure.
Therefore, even if $z_T^{\rm in}$ retains some structure from $z_0^{\rm in}$, the impact remains minimal.

\vspace{2mm}\noindent
{\bf Post-processing.}~
The RGB output $x^{\rm out}$ is transformed into the Lab color space~\cite{connolly1997study}, where post-processing is applied to both the L and ab channels.
First, the L channel of $x^{\rm out}$ is replaced with $x^{\rm in}$,
because in the colorization task, the input luminance values should be retained in the output.
Then, the ab channels of $x^{\rm out}$ are normalized to match the mean and standard deviation of the ab channels of the reference image $x^{\rm ref}$, which improves fidelity to the reference.

\vspace{2mm}\noindent
{\bf Repetition of the colorization process.}~
To improve the colorized results, we repeat the colorization process.
DDIM inversion is applied to the colorized result $z_0^{\rm out}$ to obtain $z_T^{\rm out}$, and the colorization process is repeated $N$ times.
The value of $N$ is fixed and consistent across all images.

\vspace{2mm}\noindent
{\bf Initial latent AdaIN.}~
We employ initial latent AdaIN~\cite{chung2024style}, where the standard deviation and mean of $z_T^{\rm out}$ are replaced with those of $z_T^{\rm ref}$.

\section{Experiments}
\label{sec:experiments}

\begin{figure*}[t]
    \centering
    \includegraphics[width=1\linewidth]{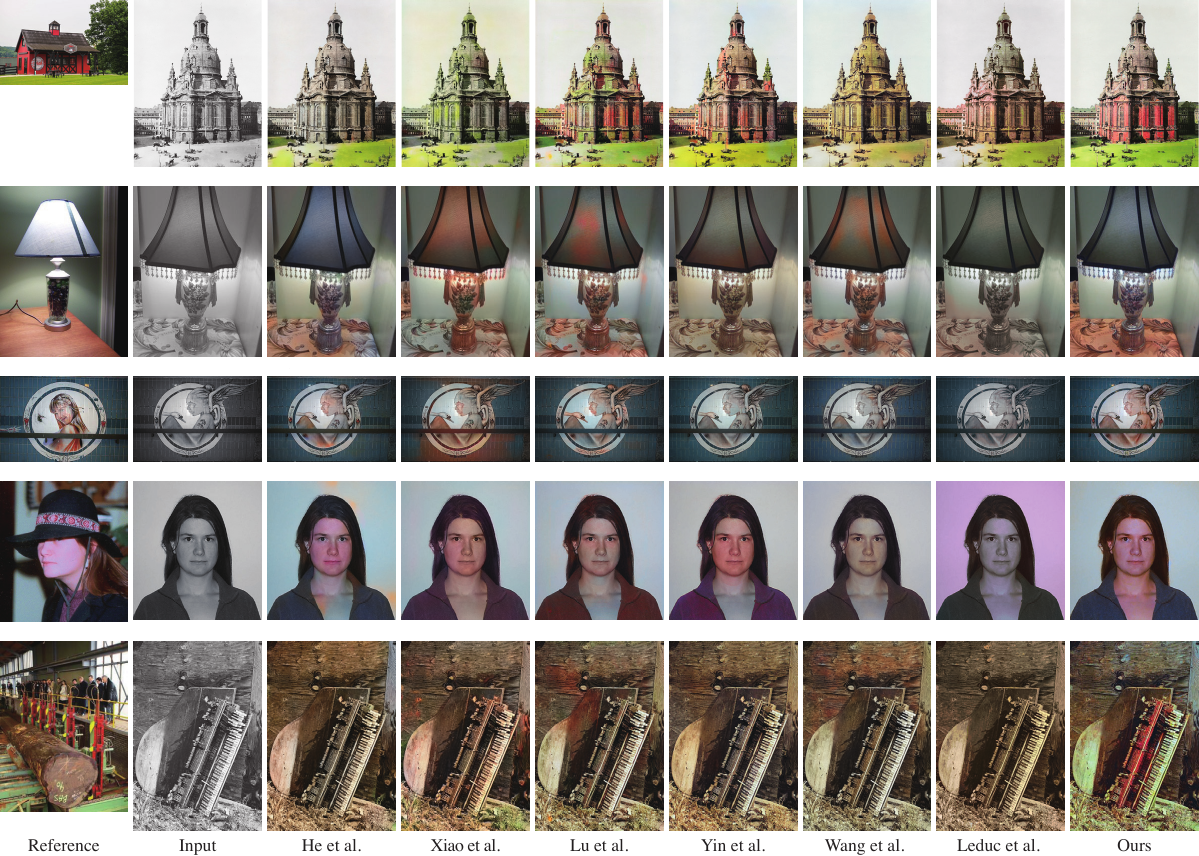}
    \vspace{-6mm}
    \caption{{Qualitative comparison.}}
    \label{fig4}
\end{figure*}

\begin{table*}[t]
  \centering
  {\footnotesize
\caption{{Quantitative comparison.} Bold and underline indicate the best and second-best scores, respectively, except in the ablation study.
\label{table1}\vspace{0mm}}
  {\tabcolsep=1.5mm
  \begin{tabular}{lccccccc}\toprule[0.4mm]
\multirow{2}{*}{\textbf{Method}} &\multicolumn{3}{c}{\textbf{Image quality}} & &\multicolumn{3}{c}{\textbf{Fidelity to the reference}}\\
\cmidrule[0.4mm]{2-4}\cmidrule[0.4mm]{6-8}
&\textbf{FID}~$\downarrow$ &\textbf{ARNIQA}~$\uparrow$ &\textbf{MUSIQ}~$\uparrow$ & &\textbf{SI-FID}~$\downarrow$ &\textbf{HIS}~$\uparrow$ &\textbf{LPIPS}~$\downarrow$\vspace{0mm}\\
\toprule[0.4mm]
Grayscale input &119.85&0.630&53.54&&11.84&0.234&0.706\vspace{0.25mm}\\
He et al.~\cite{he2018deep}&\underline{95.85}&0.666&61.75&&7.00&0.751&\underline{0.640}\vspace{0.25mm}\\
Xiao et al.~\cite{xiao2020example}&108.22&0.665&60.83&&7.06&0.709&0.651\vspace{0.25mm}\\
Lu et al.~\cite{lu2020gray2colornet}~~~~~~~~~~~&103.26&0.667&61.29&&\underline{6.11}&\underline{0.777}&0.647\vspace{0.25mm}\\
Yin et al.~\cite{yin2021yes}&102.49&{\bf 0.670}&\underline{61.94}&&6.17&0.698&0.643\vspace{0.25mm}\\
Wang et al.~\cite{wang2023unsupervised}&106.57&0.660&60.79&&7.04&0.680&0.644\vspace{0.25mm}\\
Leduc et al.~\cite{leduc2024non}&109.82&0.663&60.17&&7.04&0.651&0.643\vspace{0.25mm}\\
Ours&{\bf 95.27}&{\bf 0.670}&{\bf 62.09}&&{\bf 5.51}&{\bf 0.792}&{\bf 0.634}\vspace{0mm}\\
\midrule[0.2mm]
Ours w/ single attention (gray-to-gray)&96.58&0.669&61.83&&5.60&0.772&0.638\vspace{0.25mm}\\
Ours w/ single attention (colorized-to-color)~~~~~~~~~&96.25&0.669&61.63&&5.56&0.798&0.637\vspace{0.25mm}\\
Ours w/ single attention (gray-to-color)&100.17&0.667&60.88&&6.74&0.738&0.649\vspace{0.25mm}\\
Ours w/o classifier-free colorization guidance&99.78&0.665&61.31&&6.29&0.741&0.646\vspace{0.25mm}\\
\toprule[0.4mm]
\end{tabular}
}
}
\end{table*}

\begin{figure*}[t]
    \centering
    \vspace{3mm}
    \includegraphics[width=0.9\linewidth]{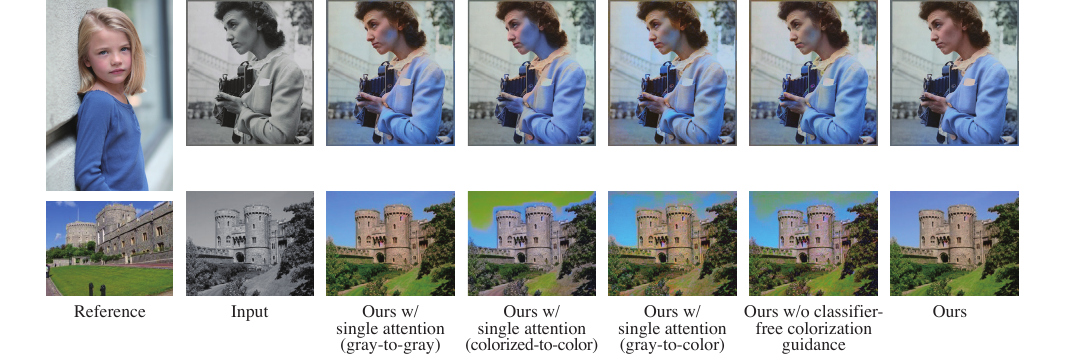}
    \vspace{-0mm}
    \caption{{Qualitative comparison in ablation studies.}}
    \label{fig5}
\end{figure*}

\subsection{Experimental Settings}
\noindent
{\bf Implementation Details.}~
We use Stable Diffusion 1.4~\cite{rombach2022high} pre-trained on the LAION dataset~\cite{schuhmann2022laion}.
We apply our method to the later layers of the decoder (the 6th to 11th layers) in the U-Net architecture.
For the deeper layers (the 6th to 8th), we set $\gamma$ to 1 and $\beta$ to 0.
For the shallower layers (the 9th to 11th), we set $\gamma$ to 0.5 and $\beta$ to 0.5.
We set $T_{\rm max}$, $T$, $w$, and $N$ to 50, 5, 10, and 3, respectively.

\vspace{2mm}
\noindent
{\bf Dataset.}~
We use 335 input-reference pairs from previous research~\cite{he2018deep}.
The images are resized to $512 \times 512$ to match the resolution on which Stable Diffusion was trained.
It should be noted that alternative resolutions can also be used for colorization, as Stable Diffusion is capable of generating images at various resolutions.
We use 8-bit images and apply bicubic interpolation for resizing. Evaluation metrics are computed based on the resized resolution.
The outputs are resized to their original sizes for visualization.

\subsection{Metrics}
In exemplar-based colorization, it is difficult to create a ground truth image,
as there is no definitive correct answer for colorizing an image based on a reference image.
Therefore, we evaluate the colorization results from two perspectives: {\bf 1) image quality} and {\bf 2) fidelity to the reference}.
To assess 1) image quality, we use three metrics.

\begin{itemize}
\vspace{1mm}
\item {\bf FID}~\cite{heusel2017gans}: the embedding distance is measured between the output image set and the reference image set.
The score range from 0 to infinity, where lower values signify better quality.
\vspace{1mm}
\item {\bf ARNIQA}~\cite{agnolucci2024arniqa}: each output is evaluated by an image quality assessment model named ARNIQA, which is trained on the FLIVE dataset~\cite{ying2020patches}.
The score ranges from 0 to 1, with higher values indicating better perceived quality.
\vspace{1mm}
\item {\bf MUSIQ}~\cite{ke2021musiq}: each output is evaluated by an image quality assessment model named MUSIQ, which is trained on the SPAQ dataset~\cite{fang2020cvpr}.
The score ranges from 0 to 100, with higher values indicating better perceived quality.
\end{itemize}

\vspace{1mm}\noindent
To assess 2) fidelity to the reference, we use three metrics.

\begin{itemize}
\vspace{1mm}
\item {\bf Single Image FID (SI-FID)}~\cite{shaham2019singan}: FID is calculated between each output and its corresponding reference individually.
The score range from 0 to infinity, where lower values signify higher fidelity.
\vspace{1mm}
\item {\bf Histogram Intersection Similarity (HIS)}~\cite{isola2017image}: the output and reference are converted to the Lab color space,
and the intersection of the two-dimensional histograms of the a and b channels is measured.
The score ranges from 0 to 1, with higher values indicating higher fidelity.
\vspace{1mm}
\item {\bf LPIPS}~\cite{zhang2018unreasonable}: perceptual similarity between each output and its corresponding reference is calculated.
The score typically ranges from 0 to 1, with lower values indicating higher fidelity.
\end{itemize}

\vspace{1mm}
As a reference, we also evaluate grayscale inputs using these metrics. In this case, the single-channel grayscale input is replicated across the channel dimension to form a three-channel image.

\subsection{Comparison with the State-of-the-Arts}
We compare our method with six existing approaches~\cite{he2018deep,xiao2020example,lu2020gray2colornet,yin2021yes,wang2023unsupervised,leduc2024non}.
Since the authors of the compared methods have released their source code, we use their original implementations for our experiments. 
We present a qualitative comparison in Fig.~\ref{fig4}.
For instance, in the reference at the top, the building walls are red and the grass on the ground is green,
and our method successfully assigns these colors to the semantically corresponding regions in the input.
Similarly, in other examples, our method captures the semantic correspondences,
resulting in better image quality and fidelity to the reference.
Quantitative comparisons are shown in Table~\ref{table1}.
Our method achieves the best score in all metrics.

The effectiveness of our method can be analyzed as follows.
The method by Xiao et al.~\cite{xiao2020example} considers only the global histogram and does not account for local semantic correspondences, 
which limits its ability to reflect the user's intent. 
The predefined feature-based approaches (He et al.~\cite{he2018deep} and Leduc et al.~\cite{leduc2024non})
utilize VGG-19~\cite{simonyan2014very}, pretrained on the ImageNet~\cite{deng2009imagenet} classification task, as a feature extractor. 
While these deep features can capture high-level information, they are not optimized for establishing local semantic correspondences, 
resulting in limited accuracy in this aspect. 
The attention-based approaches (Lu et al.\cite{lu2020gray2colornet}, Yin et al.\cite{yin2021yes}, and Wang et al.~\cite{wang2023unsupervised}) 
have a major drawback in that  training attention modules is computationally expensive.
Without a sufficiently large dataset, it is difficult to fully leverage their potential. 
In contrast, our method leverages the attention mechanism of Stable Diffusion~\cite{rombach2022high}, 
which is trained on the LAION dataset~\cite{schuhmann2022laion} containing over five billion images. 
Owing to this large-scale dataset, our approach can fully utilize the potential of attention modules, leading to more accurate semantic correspondences.

\subsection{Ablation Studies}
To demonstrate that each element of the proposed method contributes to performance improvement,
we conduct ablation studies under the following conditions.

\vspace{2mm}
\noindent
{\bf Ours w/ single attention (gray-to-gray).}
Instead of using the dual attention, we only use the gray-to-gray attention,
{\it i.e.}, we set $\gamma$ to 1 in Eq.~(\ref{dual_attention}).

\vspace{2mm}
\noindent
{\bf Ours w/ single attention (colorized-to-color).}
Instead of using the dual attention, we only use the colorized-to-color attention,
{\it i.e.}, we set $\gamma$ to 0 in Eq.~(\ref{dual_attention}).

\vspace{2mm}
\noindent
{\bf Ours w/ single attention (gray-to-color).}
Instead of using the dual attention, we only use gray-to-color attention.
During DDIM inversion, we collect the query features $q_t^{\rm in}$ from $z_t^{\rm in}$ and the key features $k_t^{\rm ref}$ from $z_t^{\rm ref}$ at each time step.
The attention map is then computed using these features as
\begin{equation}
  A_{\rm g2c} = {\rm softmax}\left(\frac{q_t^{\rm in}(k_t^{\rm ref})^T}{\sqrt{d}}\right).
\end{equation}
Then, we replace $A_{\rm dual}$ with $A_{\rm g2c}$.
This setup implies that attention is computed directly between the grayscale input and the color reference.

\vspace{2mm}
\noindent
{\bf Ours w/o classifier-free colorization guidance.}
We do not use the classifier-free colorization guidance,
{\it i.e.}, we set $w$ to 1 in Eq.~(\ref{color_enhancement}).

\vspace{2mm}
The quantitative evaluation of these settings is shown in Table~\ref{table1}, and the qualitative evaluation is shown in Fig.~\ref{fig5}.
It can be observed that the original condition outperforms the ablation settings,
indicating that the design of the dual attention-guided color transfer and classifier-free colorization guidance plays a crucial role.

\begin{figure*}[t]
    \centering
    \includegraphics[width=1\linewidth]{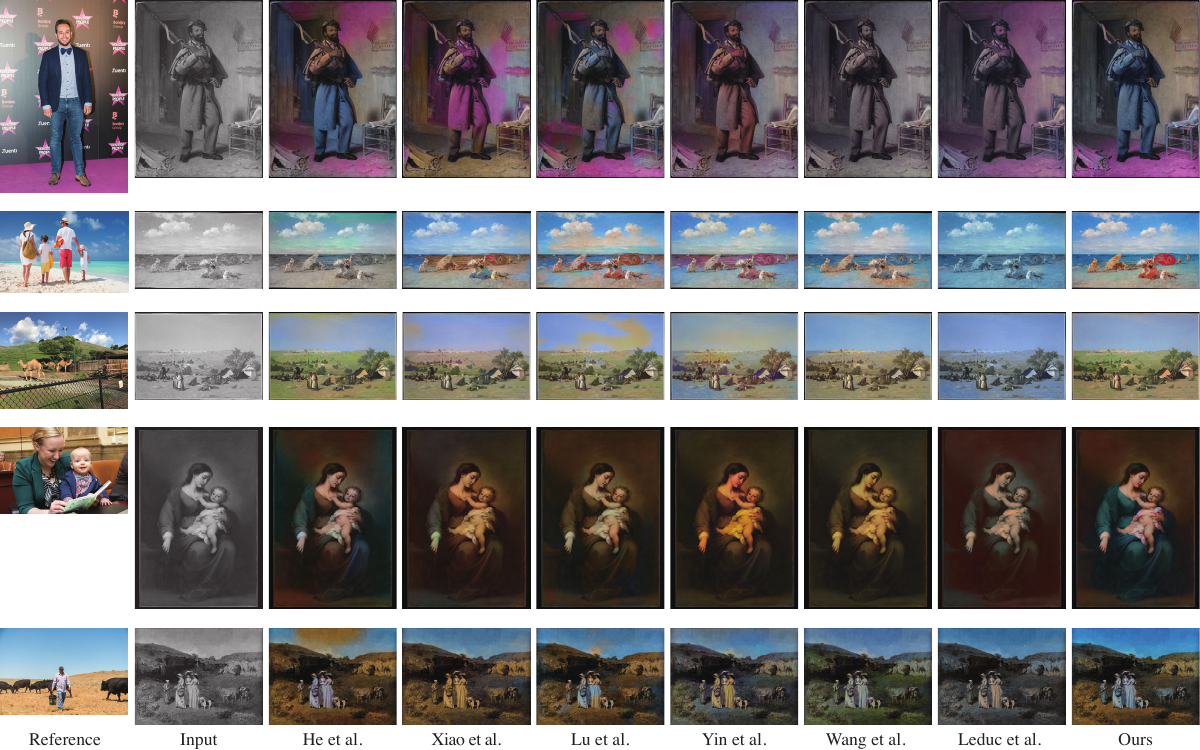}
    \vspace{-5mm}
    \caption{{Qualitative comparison in the cross-domain reference scenario.}
    The references are contemporary photos, and the inputs are historical paintings.
    }
    \label{fig6}
    \vspace{-1mm}
\end{figure*}

\begin{table}[t]
  \centering
  {\footnotesize
\caption{{Quantitative comparison in the cross-domain reference scenario.}
\label{table2}\vspace{-2.5mm}}
{\tabcolsep=0.14mm
\begin{tabular}{l@{\hspace{2.5mm}}cccc@{\hspace{1.5mm}}ccc}\toprule[0.4mm]
\multirow{2}{*}{\textbf{Method}} &\multicolumn{3}{c}{\textbf{Image quality}} & &\multicolumn{3}{c}{\textbf{Fidelity to the reference}}\\
\cmidrule[0.4mm]{2-4}\cmidrule[0.4mm]{6-8}
&\textbf{FID}$\downarrow$ &\textbf{ARNIQA}$\uparrow$ &\textbf{MUSIQ}$\uparrow$ & &\textbf{SI-FID}$\downarrow$ &\textbf{HIS}$\uparrow$ &\textbf{LPIPS}$\downarrow$\vspace{0mm}\\
\toprule[0.4mm]
Grayscale input &275.18&0.629&55.66&&13.34&0.179&0.801\vspace{0.25mm}\\
He et al.~\cite{he2018deep}&\underline{223.31}&0.664&72.31&&9.93&0.659&0.754\vspace{0.25mm}\\
Xiao et al.~\cite{xiao2020example}&236.86&0.671&71.65&&9.49&0.662&0.753\vspace{0.25mm}\\
Lu et al.~\cite{lu2020gray2colornet}&247.25&0.670&\underline{72.78}&&\underline{8.38}&\underline{0.746}&0.756\vspace{0.25mm}\\
Yin et al.~\cite{yin2021yes}&249.92&0.673&72.53&&8.43&0.606&0.760\vspace{0.25mm}\\
Wang et al.~\cite{wang2023unsupervised}&240.92&0.661&71.03&&9.17&0.604&\underline{0.748}\vspace{0.25mm}\\
Leduc et al.~\cite{leduc2024non}&244.60&\underline{0.676}&69.80&&10.29&0.540&0.755\vspace{0.25mm}\\
Ours&{\bf 219.05}&{\bf 0.681}&{\bf 72.96}&&{\bf 7.94}&{\bf 0.775}&{\bf 0.733}\vspace{0mm}\\
\toprule[0.4mm]
\end{tabular}
}
}
\end{table}

\subsection{Cross-domain Reference Scenario}
To evaluate the robustness of our method, we conduct experiments in a more challenging setting: a cross-domain reference scenario.
In this setup, we use grayscale images of historical paintings as the input and contemporary photos as the reference color images.
Historical paintings, depending on their state of preservation, may have faded or missing colors, making this a highly practical evaluation setting.

We collected 100 open-access historical paintings from the Metropolitan Museum of Art's online collection and converted them to grayscale.
Additionally, we manually gathered corresponding reference photos from Flickr, creating a dataset containing 100 input-reference pairs.

Fig.~\ref{fig6} presents a qualitative comparison.
Despite the significant domain differences between the inputs and references, our method successfully captures semantic correspondences,
thanks to the powerful attention of the pre-trained diffusion model.
Table~\ref{table2} presents a quantitative comparison, showing that our method achieves the highest performance across all metrics.
These results highlight the robustness of the proposed method.

\subsection{Comparison with Exemplar-free Methods}
\begin{figure}[t]
  \centering
  \includegraphics[width=1\linewidth]{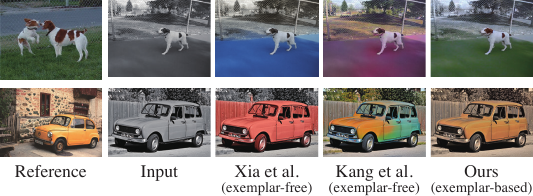}
  \vspace{-6mm}
  \caption{Qualitative comparison with exemplar-free methods.}
  \label{fig7}
  \vspace{-3mm}
\end{figure}

We compare our method with two recent exemplar-free approaches~\cite{xia2022disentangled,kang2023ddcolor}.
A quantitative comparison is presented in Table~\ref{table3}.
In terms of image quality, the difference between our method and the exemplar-free methods is relatively small.
However, our method is clearly superior in terms of fidelity to the reference.
A qualitative comparison is shown in Fig.~\ref{fig7}.
When exemplar images are not taken into account, there are multiple possible ways to colorize a grayscale image,
and exemplar-free methods do not allow users to control the colorization results.
In contrast, our method can faithfully reflect the user's intentions.

\begin{table}[t]
  \centering
  {\footnotesize
\caption{{Quantitative comparison with exemplar-free methods.}
\label{table3}\vspace{-2.5mm}}
{\tabcolsep=0.14mm
\begin{tabular}{l@{\hspace{2.5mm}}cccc@{\hspace{1.5mm}}ccc}\toprule[0.4mm]
\multirow{2}{*}{\textbf{Method}} &\multicolumn{3}{c}{\textbf{Image quality}} & &\multicolumn{3}{c}{\textbf{Fidelity to the reference}}\\
\cmidrule[0.4mm]{2-4}\cmidrule[0.4mm]{6-8}
&\textbf{FID}$\downarrow$ &\textbf{ARNIQA}$\uparrow$ &\textbf{MUSIQ}$\uparrow$ & &\textbf{SI-FID}$\downarrow$ &\textbf{HIS}$\uparrow$ &\textbf{LPIPS}$\downarrow$\vspace{0mm}\\
\toprule[0.4mm]
Xia et al.~\cite{xia2022disentangled}&106.83&0.658&{\bf 62.27}&&10.35&\underline{0.434}&0.684\vspace{0.25mm}\\
Kang et al.~\cite{kang2023ddcolor}&\underline{96.22}&\underline{0.668}&60.40&&\underline{8.49}&0.471&\underline{0.676}\vspace{0.25mm}\\
Ours&{\bf 95.27}&{\bf 0.670}&\underline{62.09}&&{\bf 5.51}&{\bf 0.792}&{\bf 0.634}\vspace{0mm}\\
\toprule[0.4mm]
\end{tabular}
}
}
\end{table}

\section{Discussion}
\subsection{Limitations}
Our method has two limitations.

\begin{figure}[t]
    \centering
    \includegraphics[width=1\linewidth]{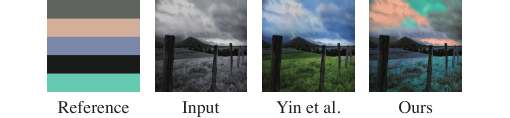}
    \vspace{-6mm}
    \caption{{Results using a color palette reference.} The reference and input are from \cite{wang2023unsupervised}.}
    \label{fig8}
    \vspace{-3mm}
\end{figure}

\vspace{2mm}
1) Since our method heavily relies on semantic correspondence, it has the limitation of not performing well with input-reference pairs that lack such correspondence.
For example, we show a result using a color palette reference in Fig.~\ref{fig8}.
Our method cannot capture the semantic correspondence between the input and the color palette, resulting in unnatural outcomes.
Future research will focus on improving our method to handle input-reference pairs that lack semantic correspondence.

\vspace{2mm}
2) Since our method repeatedly utilizes the denoising model, it has the limitation of a longer runtime compared to existing approaches.
Table~\ref{table4} presents a runtime comparison conducted using an NVIDIA RTX A6000 GPU.

One potential solution to this issue is to distill the proposed framework into a single-step denoising model.
A previous study~\cite{sauer2024adversarial} has shown that distilling a multi-step diffusion model into a single-step model can significantly accelerate inference while maintaining performance.
Similarly, distilling our colorization framework could lead to improved efficiency.

However, the primary goal of this work is to demonstrate that the attention modules of a pre-trained diffusion model can be effectively leveraged for colorization without fine-tuning.
Therefore, reducing computational cost is left as a direction for future work.

\subsection{Different Potential Applications}
While our paper primarily focuses on colorizing general images and historical paintings, image colorization has a broad range of applications including:
\begin{itemize}
\item Restoration and enhancement of old or damaged photographs and films.

\item Assisting digital artists in generating base colors for line art.

\item Improving accessibility for colorblind individuals through color enhancement.

\item Enriching training data for machine learning models by generating colored versions of grayscale images.

\item Providing visual context in medical imaging or scientific visualizations when certain features are enhanced using artificial colors.
\end{itemize}

These diverse applications highlight the broader impact and usefulness of image colorization technologies.

\begin{table}[t]
  \centering
  {\footnotesize
\caption{{Runtime comparison.}
\label{table4}\vspace{-2mm}}
  {\tabcolsep=2mm
  \begin{tabular}{lc}\toprule[0.4mm]
\textbf{Method} &\textbf{Runtime}\\
\toprule[0.4mm]
He et al.~\cite{he2018deep}&49.3 sec\vspace{0.25mm}\\
Xiao et al.~\cite{xiao2020example}&6.6 sec\vspace{0.25mm}\\
Lu et al.~\cite{lu2020gray2colornet}~~~~~~~~~~~&0.7 sec\vspace{0.25mm}\\
Yin et al.~\cite{yin2021yes}&1.1 sec\vspace{0.25mm}\\
Wang et al.~\cite{wang2023unsupervised}&0.2 sec\vspace{0.25mm}\\
Leduc et al.~\cite{leduc2024non}&23.4 sec\vspace{0.25mm}\\
Ours&5.2 sec\vspace{0.25mm}\\
\toprule[0.4mm]
\end{tabular}
}
}
\end{table}

\section{Conclusion}
\label{sec:conclusion}
We proposed a novel, fine-tuning-free exemplar-based image colorization method that leverages the self-attention mechanism of a pre-trained diffusion model. By utilizing the rich semantic representations learned from large-scale datasets, our method achieves more accurate and natural color transfer compared to prior approaches.

\vspace{2mm}
\noindent
{\bf Key Findings.}
We found that incorporating two core components significantly improves exemplar-based image colorization. 
First, our dual attention-guided color transfer—which adapts self-attention into a cross-image attention mechanism—enables 
semantically aligned colorization by leveraging both grayscale and reference color features. 
Second, classifier-free colorization guidance enhances image realism by combining color-transferred and non-color-transferred
outputs during the denoising process. 
Experiments on standard benchmarks and challenging image pairs show that our method consistently outperforms existing techniques 
in both image quality and fidelity to the reference.

\vspace{2mm}
\noindent
{\bf Potential Impact.}
This study demonstrates the effectiveness of utilizing pre-trained diffusion models for exemplar-based colorization without requiring additional training.
The proposed method holds promise for applications in historical image restoration, digital archiving, and creative media production.
Furthermore, it opens up a new direction for future research in semantically guided image translation.

\bibliographystyle{IEEEtran}
\bibliography{main}
\balance

\begin{IEEEbiography}[{\includegraphics[width=1in,height=1.25in,clip,keepaspectratio]{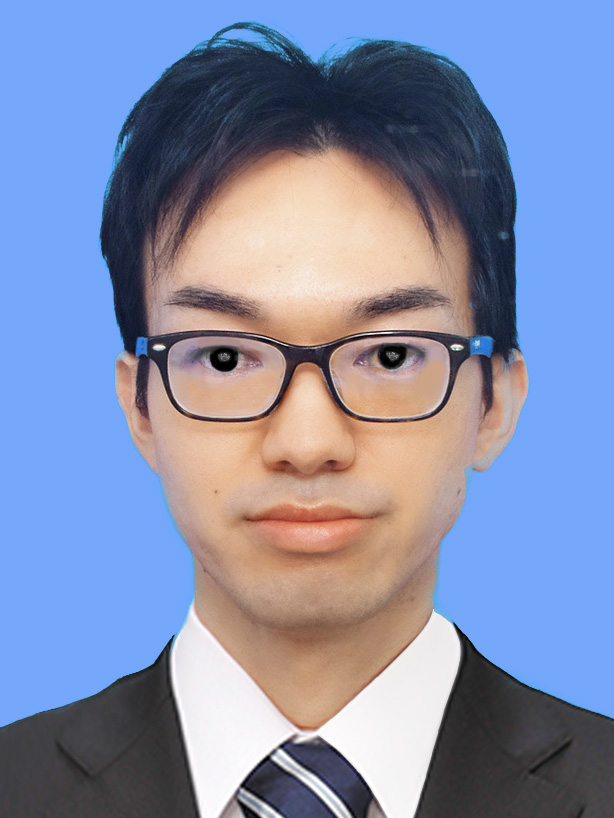}}]{Satoshi Kosugi}
  received the B.S., M.S., and Ph.D. degrees in information and communication
engineering from The University of Tokyo, Japan,
in 2018, 2020, and 2023, respectively.
He is currently an Assistant Professor at
the Laboratory for Future Interdisciplinary Research of Science and Technology,
Institute of Integrated Research, Institute of Science Tokyo.
His research interests
include computer vision, with particular interest in
image restoration.
\end{IEEEbiography}

\vfill
\clearpage

\setcounter{section}{0}
\renewcommand{\thesection}{APPENDIX \Roman{section}}

\section{Additional Ablation Studies}
We conduct additional ablation studies using the following settings.

\begin{figure*}[t]
    \centering
    \vspace{3mm}
    \includegraphics[width=1\linewidth]{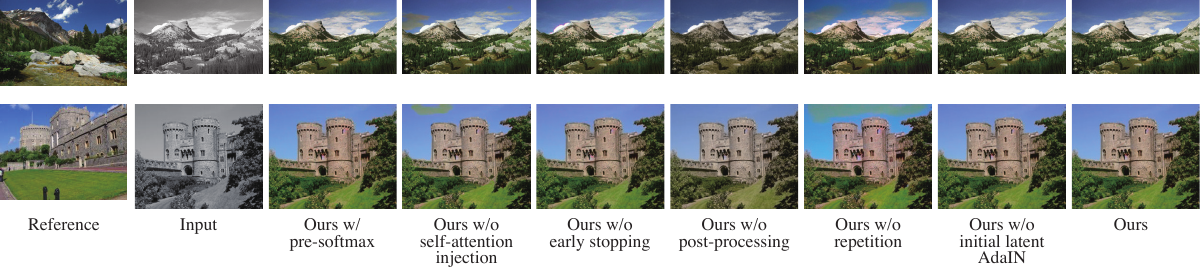}
    \vspace{-6mm}
    \caption{{Qualitative comparison in ablation studies.}}
    \label{fig9}
\end{figure*}

\vspace{2mm}
\noindent
{\bf Ours w/ pre-softmax.}
When calculating the dual attention in Eq. (10), we use a post-softmax strategy.
Specifically, we first combine the two similarity maps and then apply softmax.
In this ablation setting, we replace the post-softmax strategy with a pre-softmax strategy, where softmax is applied to create attention maps first, and then they are combined.
We modify Eq. (10) as follows.
\begin{equation}
  \begin{split}
A_{\rm g2g} &= {\rm softmax}(S_{\rm g2g}), A_{\rm c2c} = {\rm softmax}(S_{\rm c2c}), \\
&A_{\rm dual} = A_{\rm g2g} \times \gamma + A_{\rm c2c} \times (1 - \gamma).
\end{split}
\end{equation}
We show the quantitative and qualitative comparisons in Table~\ref{table5} and Fig.~\ref{fig9}, respectively.
Although the difference is small, the post-softmax strategy achieves better performance.

\vspace{2mm}
\noindent
{\bf Ours w/o self-attention injection.}
In this ablation setting, we do not use the self-attention injection,
{\it i.e.}, we set $\beta$ to 0 in Eq.~(13).
As demonstrated in Table~\ref{table5}, in some metrics, better performance is achieved when we do not use the self-attention injection.
However, as shown by the ARNIQA and MUSIQ scores in Table~\ref{table5}, as well as the qualitative comparison in Fig.~\ref{fig9},
the self-attention injection plays an important role in improving image quality.

\vspace{2mm}
\noindent
{\bf Ours w/o early stopping.}
In this ablation setting, we do not use early stopping of DDIM inversion,
{\it i.e.}, we set $T$ to $T_{\text{max}}$.
As shown in Table~\ref{table6}, employing early stopping reduces computational cost, while the performance remains largely unaffected,
as demonstrated in Table~\ref{table5} and Fig.~\ref{fig9}.

\vspace{2mm}
\noindent
{\bf Ours w/o post-processing.}
We do not apply post-processing to the ab channels.
To remain consistent with existing research, post-processing to the L channel must be applied.
As shown in Table~\ref{table5} and Fig.~\ref{fig9}, this post-processing step improves performance.

\vspace{2mm}
\noindent
{\bf Ours w/o repetition.}
In this ablation setting, we do not repeat the colorization process,
{\it i.e.}, we set $N$ to 1.
As demonstrated in Table~\ref{table5} and Fig.~\ref{fig9}, repeating the colorization process improves the results.

\vspace{2mm}
\noindent
{\bf Ours w/o initial latent AdaIN.}
In this ablation setting, we do not use initial latent AdaIN~[47].
As demonstrated in Table~\ref{table5}, the initial latent AdaIN particularly improves the FID score.

\section{Results Using Different Layers in the Decoder.}
We use only the 6th to 11th layers of the decoder in Stable Diffusion [12].
To justify this choice, we conduct experiments using different groups of decoder layers.
The decoder layers can be categorized into four groups based on depth:
the shallowest layers (9th, 10th, 11th), the second shallowest (6th, 7th, 8th), the third shallowest (3rd, 4th, 5th), and the deepest layers (0th, 1st, 2nd).
It is worth noting that no attention mechanisms are present in the deepest layers.
Table~\ref{table7} presents the results of progressively adding layer groups, starting from the shallowest up to the third shallowest.
When using the third shallowest layers, we apply the same hyperparameters as for the second shallowest group.
As shown in Table~\ref{table7}, using both the shallowest and second shallowest layers tends to yield good performance on average.

\section{Results Using Different Hyperparameters.}
To justify our hyperparameter settings, we conduct experiments using different values, especially for $\beta$, $\gamma$, $T$, $w$, and $N$.

For $\beta$ and $\gamma$, we vary them among $\{0, 0.5, 1\}$. As shown in Table~\ref{table8}, our chosen settings achieve good performance on average.

For $T$, we vary it across $\{5, 10, 50\}$. Changing $T$ does not significantly affect performance. Since a smaller $T$ leads to lower computational cost, we adopt $T=5$.

For $w$, we test values from $\{1, 5, 10, 15\}$, and the best performance is achieved when $w=10$.

For $N$, we vary it over $\{1, 2, 3, 4\}$. Although most metrics improve as $N$ increases, the FID worsens. We therefore adopt $N=3$, which achieves good overall performance.

As for $T_{\rm max}$, we fix it at 50, following previous studies~[44]-[48].

\section{Results Using the YUV Color Space}
\begin{figure*}[t]
  \centering
  \vspace{3mm}
  \includegraphics[width=1\linewidth]{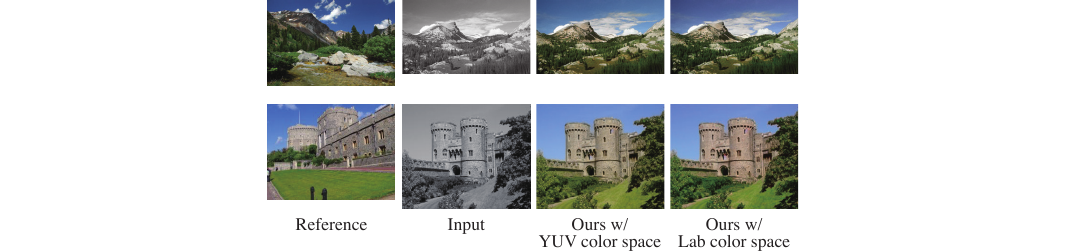}
  \vspace{-6mm}
  \caption{Qualitative comparison using different color spaces.}
  \label{fig10}
\end{figure*}

In accordance with existing methods~[14], [15], [19], [23], [40], [51], we use the Lab color space for post-processing.
However, an alternative color space, YUV, can also be used.
For completeness, we present the results using the YUV color space in Fig.~\ref{fig10} and Table~\ref{table9}.
As shown in Fig.~\ref{fig10}, the visual quality is nearly identical whether the Lab or YUV color space is used.
The quantitative scores in Table~\ref{table9} also indicate that the difference is negligible.

\section{Results Using Complex Inputs}
To evaluate the robustness of our method against complex inputs, we conduct experiments using such inputs.
We define image complexity based on the number of objects present in the image.
Specifically, we use YOLOv3~\cite{redmon2018yolov3} to detect objects in each input image and classify those with five or more objects as complex inputs. 
Among the 335 inputs used in the main paper, 16 are identified as complex inputs.

\begin{figure*}[t]
  \centering
  \vspace{3mm}
  \includegraphics[width=1\linewidth]{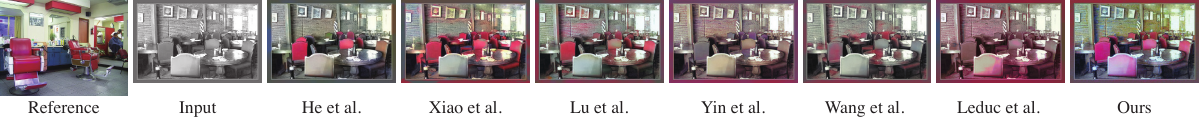}
  \vspace{-6mm}
  \caption{{Qualitative comparison using a complex input.}}
  \label{fig11}
\end{figure*}

Quantitative comparisons using the complex inputs are presented in Table~\ref{table10}, and a qualitative comparison is shown in Fig.~\ref{fig11}.
Our method demonstrates strong and consistent performance across various metrics, demonstrating its robustness to complex inputs.

\section{Quantitative Evaluation Based on Additional Image Quality Metrics}
While various image quality assessment models have been proposed, we select ARNIQA~[53] and MUSIQ~[55] for evaluation in the main paper.
This is because ARNIQA represents one of the state-of-the-art approaches, while MUSIQ is a widely adopted metric.
To provide a more comprehensive evaluation of our proposed method, we additionally assess it using the following five metrics.

\begin{itemize}
\item {\bf TOPIQ}~\cite{10478301}: each output is evaluated by an image quality assessment model named TOPIQ, which is trained on the FLIVE dataset~[54].
The score ranges from 0 to 1, with higher values indicating better perceived quality.
\vspace{1mm}

\item {\bf LIQE}~\cite{zhang2023blind}: each output is evaluated by an image quality assessment model named LIQE, which is trained on multiple datasets.
The score ranges from 1 to 5, with higher values indicating better perceived quality.
\vspace{1mm}

\item {\bf TReS}~\cite{golestaneh2022no}: each output is evaluated by an image quality assessment model named TOPIQ, which is trained on the FLIVE dataset~[54].
The score ranges from 0 to 100, with higher values indicating better perceived quality.
\vspace{1mm}

\item {\bf MANIQA}~\cite{yang2022maniqa}: each output is evaluated by an image quality assessment model named MANIQA, which is trained on the KADID dataset~\cite{lin2019kadid}.
The score ranges from 0 to 1, with higher values indicating better perceived quality.
\vspace{1mm}

\item {\bf PaQ-2-PiQ}~[54]: each output is evaluated by an image quality assessment model named PaQ-2-PiQ, which is trained on the FLIVE dataset~[54].
The score ranges from 0 to 100, with higher values indicating better perceived quality.

\end{itemize}

Furthermore, we compute the average rank based on the ranks for each metric.

Table~\ref{table11} shows the quantitative comparison based on these metrics. 
Our method achieves the best or second-best performance on almost all metrics, and has the best average rank. 
These results further support the high image quality achieved by our proposed method.

\section{Additional Qualitative Comparisons}
\begin{figure*}[t]
  \centering
  \includegraphics[width=1\linewidth]{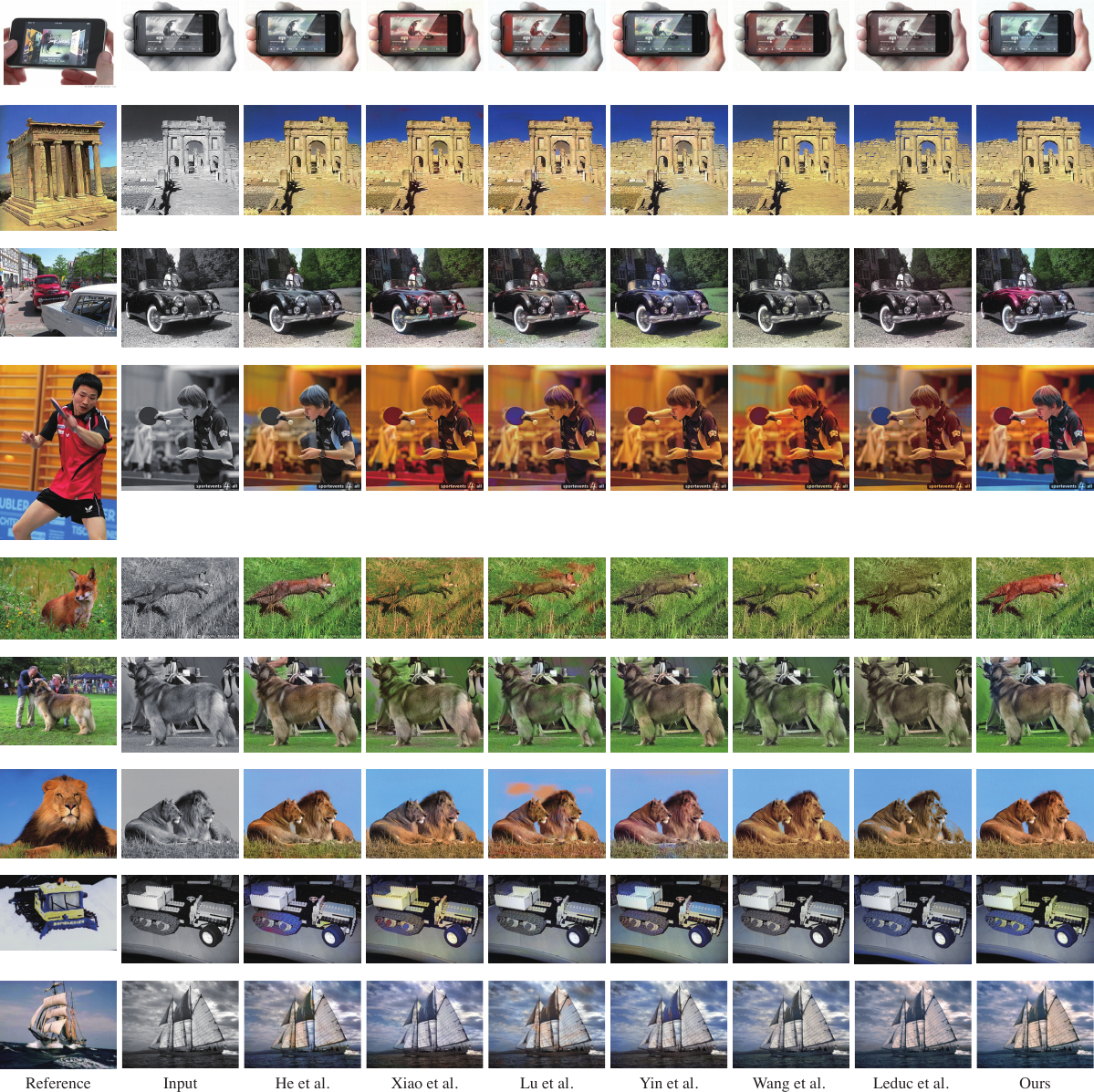}
  \vspace{-6mm}
  \caption{{Qualitative comparison.}}
  \label{fig12}
\end{figure*}

\begin{figure*}[t]
  \centering
  \includegraphics[width=1\linewidth]{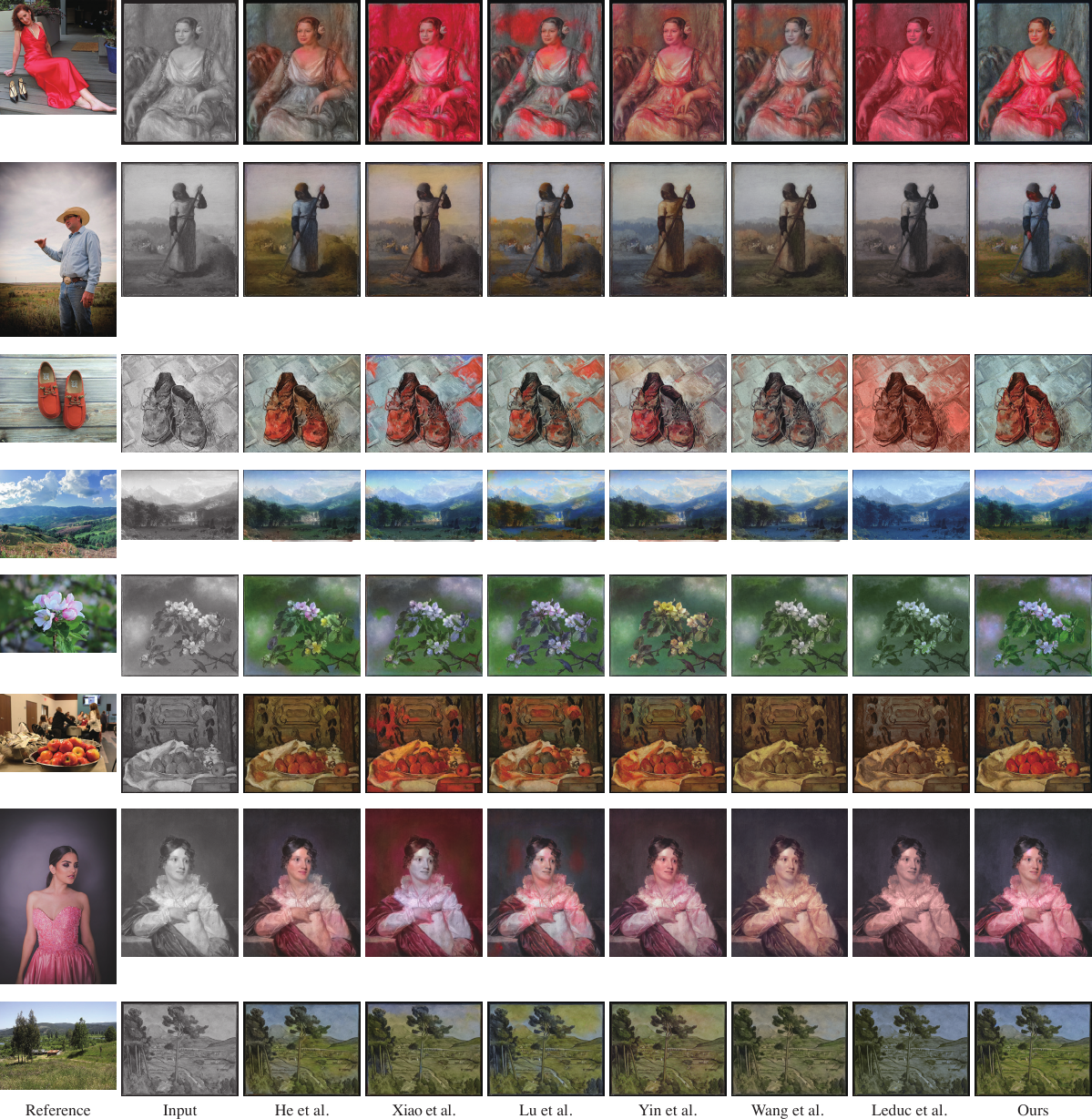}
  \vspace{-6mm}
  \caption{{Qualitative comparison in the cross-domain reference scenario.}}
  \label{fig13}
\end{figure*}

We show additional qualitative comparisons in Figs.~\ref{fig12} and \ref{fig13}.
These results further confirm that our method effectively captures semantic correspondences.

\balance

\clearpage

\begin{table*}[t]
  \centering
  {\footnotesize
\caption{{Quantitative comparison in ablation studies.}
\label{table5}\vspace{-2mm}}
  {\tabcolsep=2mm
  \begin{tabular}{lccccccc}\toprule[0.4mm]
\multirow{2}{*}{\textbf{Method}} &\multicolumn{3}{c}{\textbf{Image quality}} & &\multicolumn{3}{c}{\textbf{Fidelity to the reference}}\\
\cmidrule[0.4mm]{2-4}\cmidrule[0.4mm]{6-8}
&\textbf{FID}~$\downarrow$ &\textbf{ARNIQA}~$\uparrow$ &\textbf{MUSIQ}~$\uparrow$ & &\textbf{SI-FID}~$\downarrow$ &\textbf{HIS}~$\uparrow$ &\textbf{LPIPS}~$\downarrow$\vspace{0mm}\\
\toprule[0.4mm]
Ours&95.27&0.670&62.09&&5.51&0.792&0.634\vspace{0.25mm}\\
Ours w/ pre-softmax&95.48&0.669&61.96&&5.54&0.780&0.636\vspace{0.25mm}\\
Ours w/o self-attention injection &93.64&0.668&61.86&&5.41&0.816&0.632\vspace{0.25mm}\\
Ours w/o early stopping&95.11&0.670&62.16&&5.45&0.786&0.632\vspace{0.25mm}\\
Ours w/o post-processing&96.21&0.663&61.39&&6.63&0.675&0.635\vspace{0.25mm}\\
Ours w/o repetition&96.12&0.668&61.39&&5.48&0.774&0.637\vspace{0.25mm}\\
Ours w/o initial latent AdaIN&95.81&0.669&62.11&&5.51&0.791&0.634\vspace{0.25mm}\\
\toprule[0.4mm]
\end{tabular}
}
}
\end{table*}

\begin{table*}[t]
  \centering
  {\footnotesize
  \caption{{Comparison of computational costs with and without early stopping of DDIM Inversion.}
\label{table6}\vspace{-2mm}}
  {\tabcolsep=2mm
  \begin{tabular}{lcc}\toprule[0.4mm]
    \textbf{Method} &\textbf{Runtime}&\textbf{GPU memory}\\
\toprule[0.4mm]
Ours&5.2 sec&14 GB\vspace{0.25mm}\\
Ours w/o early stopping&36.7 sec&31 GB\vspace{0.25mm}\\
\toprule[0.4mm]
\end{tabular}
}
}
\end{table*}

\clearpage

\begin{table*}[t]
  \centering
  {\footnotesize
\caption{{Quantitative comparison using different layers in the decoder. The gray fill denotes the adopted setting.}
\label{table7}\vspace{-2mm}}
  {\tabcolsep=1mm
  \begin{tabular}{cccccccccc}\toprule[0.4mm]
    \multirow{2}{*}{\begin{tabular}{c}{\textbf{The shallowest}}\\{\textbf{layers (9th, 10th, 11th)}\vspace{-1mm}}\end{tabular}} &\multirow{2}{*}{\begin{tabular}{c}{\textbf{The second shallowest}}\\{\textbf{layers (6th, 7th, 8th)}\vspace{-1mm}}\end{tabular}}&\multirow{2}{*}{\begin{tabular}{c}{\textbf{The third shallowest}}\\{\textbf{layers (3rd, 4th, 5th)}\vspace{-1mm}}\end{tabular}} &\multicolumn{3}{c}{\textbf{Image quality}} & &\multicolumn{3}{c}{\textbf{Fidelity to the reference}}\\
\cmidrule[0.4mm]{4-6}\cmidrule[0.4mm]{8-10}
 & & &\textbf{FID}~$\downarrow$ &\textbf{ARNIQA}~$\uparrow$ &\textbf{MUSIQ}~$\uparrow$ & &\textbf{SI-FID}~$\downarrow$ &\textbf{HIS}~$\uparrow$ &\textbf{LPIPS}~$\downarrow$\vspace{0mm}\\
\toprule[0.4mm]
$\checkmark$&&&96.24&0.668&61.78&&5.79&0.789&0.637\vspace{0.25mm}\\
\rowcolor{gray!20}
$\checkmark$&$\checkmark$&&95.27&0.670&62.09&&5.51&0.792&0.634\vspace{0.25mm}\\
$\checkmark$&$\checkmark$&$\checkmark$&95.37&0.669&62.09&&5.50&0.792&0.634\vspace{0.25mm}\\
\toprule[0.4mm]
\end{tabular}
}
}
\end{table*}

\newpage

\begin{table*}[t]
  \centering
  {\footnotesize
\caption{{Quantitative comparison using different hyperparameters. The gray fill denotes the adopted setting.}
\label{table8}\vspace{-2mm}}
  {\tabcolsep=2mm
  \begin{tabular}{lccccccc}\toprule[0.4mm]
\multirow{2}{*}{\textbf{Parameter}} &\multicolumn{3}{c}{\textbf{Image quality}} & &\multicolumn{3}{c}{\textbf{Fidelity to the reference}}\\
\cmidrule[0.4mm]{2-4}\cmidrule[0.4mm]{6-8}
&\textbf{FID}~$\downarrow$ &\textbf{ARNIQA}~$\uparrow$ &\textbf{MUSIQ}~$\uparrow$ & &\textbf{SI-FID}~$\downarrow$ &\textbf{HIS}~$\uparrow$ &\textbf{LPIPS}~$\downarrow$\vspace{0mm}\\
\toprule[0.4mm]
$\beta$ (for the 9th, 10th, 11th layers) $=0$&93.53&0.668&61.87&&5.41&0.816&0.632\vspace{0.25mm}\\
\rowcolor{gray!20}
$\beta$ (for the 9th, 10th, 11th layers) $=0.5$&95.27&0.670&62.09&&5.51&0.792&0.634\vspace{0.25mm}\\
$\beta$ (for the 9th, 10th, 11th layers) $=1$&98.53&0.670&61.64&&5.64&0.732&0.640\vspace{0.25mm}\\
\midrule[0.2mm]
\rowcolor{gray!20}
$\beta$ (for the 6th, 7th, 8th layers) $=0$&95.27&0.670&62.09&&5.51&0.792&0.634\vspace{0.25mm}\\
$\beta$ (for the 6th, 7th, 8th layers) $=0.5$&95.72&0.670&62.05&&5.64&0.781&0.636\vspace{0.25mm}\\
$\beta$ (for the 6th, 7th, 8th layers) $=1$&97.30&0.669&61.95&&5.77&0.773&0.638\vspace{0.25mm}\\
\midrule[0.2mm]
$\gamma$ (for the 9th, 10th, 11th layers) $=0$&95.16&0.669&61.97&&5.49&0.794&0.634\vspace{0.25mm}\\
\rowcolor{gray!20}
$\gamma$ (for the 9th, 10th, 11th layers) $=0.5$&95.27&0.670&62.09&&5.51&0.792&0.634\vspace{0.25mm}\\
$\gamma$ (for the 9th, 10th, 11th layers) $=1$&96.58&0.669&61.83&&5.60&0.772&0.638\vspace{0.25mm}\\
\midrule[0.2mm]
$\gamma$ (for the 6th, 7th, 8th layers) $=0$&94.98&0.670&61.87&&5.53&0.801&0.634\vspace{0.25mm}\\
$\gamma$ (for the 6th, 7th, 8th layers) $=0.5$&94.85&0.669&62.00&&5.53&0.800&0.634\vspace{0.25mm}\\
\rowcolor{gray!20}
$\gamma$ (for the 6th, 7th, 8th layers) $=1$&95.27&0.670&62.09&&5.51&0.792&0.634\vspace{0.25mm}\\
\midrule[0.2mm]
\rowcolor{gray!20}
$T=5$&95.27&0.670&62.09&&5.51&0.792&0.634\vspace{0.25mm}\\
$T=10$&95.35&0.670&62.15&&5.53&0.789&0.634\vspace{0.25mm}\\
$T=50$&95.11&0.670&62.16&&5.45&0.786&0.632\vspace{0.25mm}\\
\midrule[0.2mm]
$w=1$&99.78&0.665&61.31&&6.29&0.741&0.646\vspace{0.25mm}\\
$w=5$&95.53&0.670&61.97&&5.62&0.781&0.635\vspace{0.25mm}\\
\rowcolor{gray!20}
$w=10$&95.27&0.670&62.09&&5.51&0.792&0.634\vspace{0.25mm}\\
$w=15$&96.02&0.667&61.70&&5.59&0.787&0.639\vspace{0.25mm}\\
\midrule[0.2mm]
$N=1$&96.12&0.668&61.39&&5.48&0.774&0.637\vspace{0.25mm}\\
$N=2$&95.12&0.669&62.02&&5.50&0.789&0.635\vspace{0.25mm}\\
\rowcolor{gray!20}
$N=3$&95.27&0.670&62.09&&5.51&0.792&0.634\vspace{0.25mm}\\
$N=4$&95.70&0.670&62.11&&5.51&0.793&0.634\vspace{0.25mm}\\
\toprule[0.4mm]
\end{tabular}
}
}
\end{table*}

\begin{table*}[h]
  \centering
  {\footnotesize
\caption{{Quantitative comparison using different color spaces.}
\label{table9}\vspace{-2mm}}
  {\tabcolsep=2mm
  \begin{tabular}{lccccccc}\toprule[0.4mm]
\multirow{2}{*}{\textbf{Method}} &\multicolumn{3}{c}{\textbf{Image quality}} & &\multicolumn{3}{c}{\textbf{Fidelity to the reference}}\\
\cmidrule[0.4mm]{2-4}\cmidrule[0.4mm]{6-8}
&\textbf{FID}~$\downarrow$ &\textbf{ARNIQA}~$\uparrow$ &\textbf{MUSIQ}~$\uparrow$ & &\textbf{SI-FID}~$\downarrow$ &\textbf{HIS}~$\uparrow$ &\textbf{LPIPS}~$\downarrow$\vspace{0mm}\\
\toprule[0.4mm]
Ours w/ Lab color space&95.27&0.670&62.09&&5.51&0.792&0.634\vspace{0.25mm}\\
Ours w/ YUV color space&95.41&0.670&62.01&&5.50&0.786&0.634\vspace{0.25mm}\\
\toprule[0.4mm]
\end{tabular}
}
}
\end{table*}

\begin{table*}[t]
  \centering
  {\footnotesize
\caption{{Quantitative comparison using complex inputs.}
\label{table10}\vspace{0mm}}
  {\tabcolsep=1.5mm
  \begin{tabular}{lccccccc}\toprule[0.4mm]
\multirow{2}{*}{\textbf{Method}} &\multicolumn{3}{c}{\textbf{Image quality}} & &\multicolumn{3}{c}{\textbf{Fidelity to the reference}}\\
\cmidrule[0.4mm]{2-4}\cmidrule[0.4mm]{6-8}
&\textbf{FID}~$\downarrow$ &\textbf{ARNIQA}~$\uparrow$ &\textbf{MUSIQ}~$\uparrow$ & &\textbf{SI-FID}~$\downarrow$ &\textbf{HIS}~$\uparrow$ &\textbf{LPIPS}~$\downarrow$\vspace{0mm}\\
\toprule[0.4mm]
Grayscale input &275.88&0.615&52.28&&11.50&0.243&0.713\vspace{0.25mm}\\
He et al.~[23]&\underline{228.84}&0.659&{\bf 61.15}&&7.19&0.726&\underline{0.663}\vspace{0.25mm}\\
Xiao et al.~[51]&239.96&0.660&58.62&&7.28&0.661&0.673\vspace{0.25mm}\\
Lu et al.~[14]~~~~~~~~~~~&240.85&0.664&60.15&&{\bf 5.92}&\underline{0.755}&0.673\vspace{0.25mm}\\
Yin et al.~[15]&229.82&{\bf 0.666}&59.23&&{\bf 5.92}&0.649&0.665\vspace{0.25mm}\\
Wang et al.~[19]&232.37&0.650&59.38&&6.86&0.631&0.668\vspace{0.25mm}\\
Leduc et al.~[40]&255.77&0.663&58.04&&7.09&0.583&0.669\vspace{0.25mm}\\
Ours&{\bf 224.28}&{\bf 0.666}&\underline{61.13}&&5.98&{\bf 0.761}&{\bf 0.662}\vspace{0mm}\\
\toprule[0.4mm]
\end{tabular}
}
}
\end{table*}

\begin{table*}[t]
  \centering
  {\footnotesize
\caption{{Quantitative comparison based on additional image quality metrics.}
\label{table11}\vspace{-2.5mm}}
{\tabcolsep=1.5mm
\begin{tabular}{l@{\hspace{5mm}}ccccccccc}
  \multicolumn{10}{c}{\textbf{(a) Standard scenario}\vspace{0.25mm}} \\\toprule[0.4mm]
\multirow{2}{*}{\textbf{Method}} &\multicolumn{8}{c}{\textbf{Image quality}} \\
\cmidrule[0.4mm]{2-10}
&\textbf{FID}~$\downarrow$ &\textbf{ARNIQA}~$\uparrow$ &\textbf{MUSIQ}~$\uparrow$ & \textbf{TOPIQ}~$\uparrow$ &\textbf{LIQE}~$\uparrow$ &\textbf{TReS}~$\uparrow$ & \textbf{MANIQA}~$\uparrow$ &\textbf{PaQ-2-PiQ}~$\uparrow$ &\textbf{Average Rank}~$\downarrow$ \vspace{0mm}\\
\toprule[0.4mm]
Grayscale input&119.85&0.630&53.54&0.721&2.14&81.82&0.282&67.90&7.875\vspace{0.25mm}\\
He et al.~[23]&\underline{95.85}&0.666&61.75&{\bf 0.749}&{\bf 2.71}&86.86&0.407&71.63&\underline{2.75}\vspace{0.25mm}\\
Xiao et al.~[51]&108.22&0.665&60.83&{\bf 0.749}&2.40&86.26&{\bf 0.517}&71.22&4.25\vspace{0.25mm}\\
Lu et al.~[14]&103.26&0.667&61.29&0.737&2.46&{\bf 87.34}&0.366&{\bf 72.11}&3.375\vspace{0.25mm}\\
Yin et al.~[15]&102.49&{\bf 0.670}&\underline{61.94}&0.740&2.54&86.88&0.377&71.84&2.875\vspace{0.25mm}\\
Wang et al.~[19]&106.57&0.660&60.79&0.735&2.37&85.56&0.328&70.86&6\vspace{0.25mm}\\
Leduc et al.~[40]&109.82&0.663&60.17&0.727&2.11&85.21&0.297&70.64&7\vspace{0.25mm}\\
Ours&{\bf 95.27}&{\bf 0.670}&{\bf 62.09}&0.746&\underline{2.65}&\underline{87.11}&\underline{0.489}&\underline{72.05}&{\bf 1.75}\vspace{0.25mm}\\
\toprule[0.4mm]
\multicolumn{10}{c}{\vspace{-2mm}}\\
\multicolumn{10}{c}{\textbf{(b) Cross-domain reference scenario}\vspace{0.25mm}}\\
\toprule[0.4mm]
Grayscale input&275.18&0.629&55.66&0.648&2.29&81.32&0.258&67.06&8\vspace{0.25mm}\\
He et al.~[23]&\underline{223.31}&0.664&72.31&0.700&2.87&86.83&0.442&71.51&4\vspace{0.25mm}\\
Xiao et al.~[51]&236.86&0.671&71.65&0.698&2.57&86.59&{\bf 0.643}&71.87&4\vspace{0.25mm}\\
Lu et al.~[14]&247.25&0.670&\underline{72.78}&\underline{0.703}&{\bf 2.89}&{\bf 88.22}&0.438&\underline{72.67}&\underline{3}\vspace{0.25mm}\\
Yin et al.~[15]&249.92&0.673&72.53&0.696&2.78&87.56&0.444&72.01&3.875\vspace{0.25mm}\\
Wang et al.~[19]&240.92&0.661&71.03&0.682&2.61&86.00&0.331&71.15&6.125\vspace{0.25mm}\\
Leduc et al.~[40]&244.60&\underline{0.676}&69.80&0.684&2.39&86.33&0.318&71.52&5.625\vspace{0.25mm}\\
Ours&{\bf 219.05}&{\bf 0.681}&{\bf 72.96}&{\bf 0.707}&\underline{2.88}&\underline{88.08}&\underline{0.573}&{\bf 72.88}&{\bf 1.375}\vspace{0.25mm}\\
\toprule[0.4mm]
\end{tabular}
}
}
\end{table*}

\end{document}